\documentclass[conference]{IEEEtran}
\usepackage{eurosym}
\def\delequal{\mathrel{\ensurestackMath{\stackon[1pt]{=}{\scriptstyle\Delta}}}}
\usepackage{balance}
\usepackage{amsmath}
\DeclareMathOperator*{\argmax}{arg\,max}

\usepackage{balance}
\usepackage{lastpage}
    
\usepackage{xcolor}
\def\BibTeX{{\rm B\kern-.05em{\sc i\kern-.025em b}\kern-.08em
    T\kern-.1667em\lower.7ex\hbox{E}\kern-.125emX}}
\usepackage{epsfig}
\usepackage{amsmath}
\usepackage{theorem}
\usepackage{mathrsfs}
\usepackage{amsfonts}
\usepackage{stackengine}
\usepackage{tikz}
\usepackage{pdfrender}
\usepackage{subfigure}
\usepackage{caption}
\usepackage{csquotes}
\usepackage{subfigure}
\usepackage{tabularx}
\usepackage{caption}
\usepackage{csquotes}
\usepackage{multirow}
\usepackage{amssymb}
\usepackage{mathrsfs}
\usepackage{euscript}
\usepackage{graphicx}
\usepackage{multirow}
\usepackage{xcolor}
\usepackage{algorithmicx}
\usepackage{algorithm}
\usepackage{algpseudocode}
\usepackage{cite}
\usepackage{url}
\usepackage{amsmath}
\usepackage[export]{adjustbox}
\usepackage{algorithm}
\usepackage{amssymb}
\usepackage{mathrsfs}
\usepackage{euscript}
\usepackage{graphicx}
\usepackage{multirow}
\usepackage{xcolor}
\usepackage{algorithmicx}
\usepackage{algorithm}
\usepackage{algpseudocode}
\usepackage{url}
\usepackage{amsmath}
\usepackage[export]{adjustbox}
\usepackage{algorithm}
\usepackage{amsmath}
\usepackage{amsfonts}
\usepackage{amssymb}
\usepackage{graphicx}
\usepackage{epstopdf}%

\providecommand{\U}[1]{\protect\rule{.1in}{.1in}}

\begin{document}

\title{Liquid State Machine-Empowered Reflection Tracking in RIS-Aided THz Communications\vspace{-0.50cm}}
\author{\IEEEauthorblockN 
{Hosein~Zarini$^{\dag}$, Narges Gholipoor$^{\star}$, Mohamad Robat Mili$^{\S}$, Mehdi~Rasti$^{\dag}$, Hina Tabassum$^{\dag\dag}$ and Ekram Hossain$^{\star\star}$
}
\vspace{-0.30cm}
		\\$^{\dag}$Dept. of Computer Engineering, Amirkabir University of Technology, Tehran, Iran\\
		$^\star$Dept. of Electrical and Computer Engineering,        Tarbiat Modares University, Tehran, Iran\\
        $^\S$Pasargad Institute for Advanced Innovative Solutions (PIAIS), Tehran, Iran\\
        $^{\dag\dag}$Dept. of Electrical Engineering and Computer Science, York University, Canada\\
        $^{\star\star}$Dept. of Electrical and Computer Engineering, University of Manitoba, Canada
	}
\maketitle\vspace{-0.20cm}
\begin{abstract}
Passive beamforming in reconfigurable intelligent surfaces (RISs) enables a feasible and efficient way of communication when the RIS reflection coefficients are precisely adjusted. In this paper, we present a framework to track the RIS reflection coefficients with the aid of deep learning from a time-series prediction perspective in a terahertz (THz) communication system. The proposed framework achieves a two-step enhancement over the similar learning-driven counterparts. Specifically, in the first step, we train a liquid state machine (LSM) to track the historical RIS reflection coefficients at prior time steps (known as a time-series sequence) and predict their upcoming time steps. We also fine-tune the trained LSM through Xavier initialization technique to decrease the prediction variance, thus resulting in a higher prediction accuracy. In the second step, we use ensemble learning technique which leverages on the prediction power of multiple LSMs to minimize the prediction variance and improve the precision of the first step. It is numerically demonstrated that, in the first step, employing the Xavier initialization technique to fine-tune the LSM results in at most 26\% lower LSM prediction variance and as much as 46\% achievable spectral efficiency (SE) improvement over the existing counterparts, when an RIS of size 11$\times$11 is deployed. In the second step, under the same computational complexity of training a single LSM, the ensemble learning with multiple LSMs degrades the prediction variance of a single LSM up to 66\% and improves the system achievable SE at most 54\%.
\end{abstract}
\begin{IEEEkeywords}
reconfigurable intelligent surface (RIS), terahertz (THz), liquid state machine (LSM), Xavier initialization technique, ensemble learning technique\vspace{-0.20cm}
\end{IEEEkeywords}
\IEEEpeerreviewmaketitle
\section{Introduction}
\par Terahertz (THz) communications have been envisioned as an emerging technology that can offer multi-Giga-bit per second (Gbps) transmissions.  Despite of various advantages, THz communications suffer from extreme attenuation and dynamic blockages leading to  short-range  communications. In this context, reconfigurable intelligent  surface (RIS)-aided techniques can be applied to compensate for the THz attenuation and blockages\cite{IRSDEF2}.  Architecturally, a typical RIS includes a number of low cost, power-efficient, programmable and configurable passive elements, together with the printed dipoles to reflect the incident radio frequency waves into intended directions. Moreover, RISs do not require  radio frequency chains (RF chains)\footnote{The components which considerably increase the hardware cost and energy consumption.} for reflecting the incident signals. In this sense, adjusting the RIS reflection coefficients (also called passive beamforming) for signal reflection is far more feasible and efficient compared to active beamforming design in THz band with a large number of RF chains at base stations (BSs). In practice, RIS-assisted systems have the potential to achieve energy and spectral efficient communications while combating the attenuation in THz bands\cite{9120206}. 

Reviewing literature reveals that designing the RIS reflection coefficients is computationally expensive through the existing optimization-driven frameworks \cite{9219206}. In the latest research works corresponding to RIS reflection coefficients design, machine/deep learning techniques have been applied to relieve this computational burden. For instance, the authors of \cite{DRL} investigated deep reinforcement learning (DRL) to this end. Unfortunately, the complexity order of action space in DRL escalates exponentially with the increasing number of RIS elements, thus making this method not scalable. In \cite{LSTM}, the authors employed a long-short-term-memory (LSTM) for configuring the RIS reflection coefficients. Due to fixed and non-optimized training parameters in LSTM, the proposed scheme in \cite{LSTM} is prone to overfitting\footnote{Overfitting by definition is the poor prediction power and the generalization capability of learning networks.} phenomenon for large-scale RIS systems. 
\par In this paper, we propose a reflection coefficients tracking framework in an RIS-aided THz system from a time-series prediction point of view with a two-step improvement over the existing learning-driven literature such as \cite{DRL} and \cite{LSTM}. 
The main contributions of this paper can be outlined as follows:
\begin{itemize}
\item In the first step of the proposed framework, a liquid state machine (LSM) \cite{schrauwen2007overview} is trained by deepMIMO dataset\cite{DeepMIMO} to track the
historical RIS reflection coefficients in time-series form and predict their upcoming time steps. We further fine-tune the trained LSM by Xavier initialization\cite{Xavier} technique. In doing so, we derive the LSM initial input weights from a zero mean Gaussian distribution with a finite variance. As a result, the LSM prediction variance falls notably, whereas its prediction accuracy improves. As the result, the achievable spectral efficiency (SE) in the RIS-assisted THz system remarkably improves.
\item We further enhance the RIS reflection coefficients tracking performance in the second step by means of ensemble learning by incorporating multitude of Xavier-enabled LSMs, each referred to as a weak learner. By applying a bootstrap aggregation mechanism as an efficient ensemble learning over the weak learners, a strong learner is ultimately developed with an excellent prediction accuracy.
\item Through analyzing standard deviation and mean absolute deviation, 17\% and 26\% lower prediction variance are observed respectively, on account of fine-tuning the LSM by means of Xavier initialization technique. Thanks to the first step, the total achievable SE in the RIS-assisted THz system outperforms the DRL \cite{DRL} and the LSTM \cite{LSTM} schemes by 46\% and 28\%, respectively, for an RIS of size 11$\times$11. While keeping the same computational complexity of training a single LSM, the ensemble learning technique by employing multitude of LSMs in the second step achieves 52\% and 66\% lower prediction variance for the standard deviation and mean absolute deviation metrics, respectively. Simulation results show that for an RIS of size 11$\times$11, the ensembling model outperforms the achievable SE in \cite{DRL} and \cite{LSTM} at most 54\% and 39\%, respectively. 
\end{itemize}
\par The remainder of this paper is organized as follows. Section II describes a system setup for the RIS-aided THz system, wherein the problem of tracking the RIS reflection coefficients is formally stated. In Section III, time-series prediction, LSM architecture, Xavier initialization and ensemble learning techniques are briefly elaborated. Complexity analysis, simulation results and conclusions are finally presented in Sections IV, V and VI, respectively.


\section{System Model and Assumptions}%

\begin{figure}
\centering
\includegraphics[width=8.0cm,height=7cm]{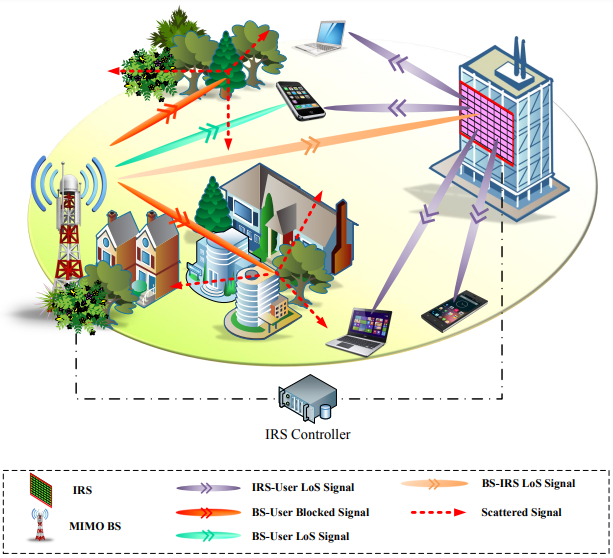}%
\caption{An RIS-assisted wireless communication system.}%
\label{Def}%
\vspace{-0.30cm}

\end{figure}

Consider a BS equipped with $N_{t}$ transmit antennas to serve $K$
single-antenna users in downlink wideband THz communication using $S$ subcarriers. Due to the short-range coverage of the THz bands, the communicating network is assisted by an RIS with $M$ passive elements and a controller module to reconfigure phase shifts of the RIS reflecting elements, deployed on LoS of the BS as shown in Fig. \ref{Def}. 
Similar to\cite{DRL} and \cite{LSTM}, it is assumed that the perfect channel state information (CSI) is available at both the BS and the RIS.
We denote the channel matrix from the BS to the RIS over the subcarrier $s$ by $\mathbf{G}[s]\in
\mathbb{C}^{M\times N_{t}}$. For a clustered ray-based wideband THz channel, according to the well-known Saleh-Valenzuela geometric model \cite{Saleh}, $\mathbf{G}[s]$ is given by  
	$\mathbf{G}[s]=
	\sum_{d=0}^{D-1}\sum_{{l}=1}^{N^{\{1\}}_{\mathrm{cl}}} \sum_{{u}=1}^{N^{\{1\}}_{\mathrm{ray}}} \alpha _{{l,u}}\delta (d T-\tau_{l,u})\Psi[s]$,
with $N^{\{1\}}_{\mathrm{cl}}$ clusters, $N^{\{1\}}_{\mathrm{ray}}$ rays and the complex gain $\alpha$. Besides, $\Psi[s]=\mathbf{a}_{\textrm{RIS}}\left(\phi_{\textrm{RIS}}^{{l,u}}[s]\right)\mathbf{a}^{H}_{\textrm{BS}}\left(\phi_{\textrm{BS}}^{{l,u}}[s]\right)e^{-j 2\pi d\frac{s}{S}}$, $j$ denotes the imaginary unit, $\delta$ is the band-limited pulse-shaping filter, $T$ is the cycle prefix length and $D$ is the sampling time. Moreover, $\mathbf{a}_{\textrm{BS}}\left(\phi\right)\in{\mathbb{C}^{N\times1}}$ and $\mathbf{a}_{\textrm{RIS}}\left(\phi\right)\in{\mathbb{C}^{M\times1}}$ are the antenna array response vectors employing uniform linear array (ULA) at the BS and the RIS, respectively and given by
$\mathbf{a}_{\textrm{BS}}(\phi)= \frac{1}{\sqrt{N}_t}\bigg[1,e^{-j 2\pi \phi_{\textrm{BS}}^{l,u}[s]},...,e^{-j 2\pi (N-1)\phi_{\textrm{BS}}^{l,u}[s]}\bigg],$
and
$\mathbf{a}_{\textrm{RIS}}(\phi)= \frac{1}{\sqrt{M}}\bigg[1,e^{-j 2\pi \phi_{\textrm{RIS}}^{l,u}[s]},...,e^{-j 2\pi (M-1)\phi_{\textrm{RIS}}^{l,u}[s]}\bigg],$
wherein $\phi_{\textrm{BS}}$ ($\phi_{\textrm{RIS}}$) are the spatial angle-of-departure (AoD) directions for the BS (RIS), defined as $\phi_{\textrm{BS}}=\phi_{\textrm{RIS}}=\frac{b}{\lambda[s]}$, with antenna spacing $b$ and wavelength of the subcarrier $s$, denoted by $\lambda[s]$.
We also denote by $\mathbf{h}_{d,k}[s]\in
\mathbb{C}
^{N_{t}\times1}$ and $\mathbf{h}_{r,k}[s]\in
\mathbb{C}
^{M\times1}$, the direct channel from the BS to the user $k$ and the reflecting channel from the RIS to this user, both over the subcarrier $s$, respectively and expressed as 

\vspace{-0.30cm}
\begin{small}
\begin{align} \label{channel2}
	\!\!h_{d,k}[s]\!=
	\!\!\!\!\sum_{d=0}^{D-1}\sum_{{l}=1}^{N^{\{2\}}_{\mathrm{cl}}} \sum_{{u}=1}^{N^{\{2\}}_{\mathrm{ray}}}\! \alpha _{{l,u}}\delta (d T\!-\!\tau_{l,u})\mathbf{a}_{\textrm{BS}}\!\left(\Bar{\phi}_{\textrm{BS},k}^{{l,u}}[s]\right)\!e^{-j 2\pi d\frac{s}{S}},
\end{align}
\end{small}
\!and
\vspace{-0.30cm}
\begin{small}
\begin{align} \label{channel3}
	\!\!\!\!h_{r,k}[s]\!=
	\!\!\!\!\sum_{d=0}^{D-1}\sum_{{l}=1}^{N^{\{3\}}_{\mathrm{cl}}} \!\sum_{{u}=1}^{N^{\{3\}}_{\mathrm{ray}}}\! \!\alpha _{{l,u}}\delta (d T\!\!-\!\tau_{l,u})\mathbf{a}_{\textrm{RIS}}\!\left(\Bar{\phi}_{\textrm{RIS},k}^{{l,u}}[s]\right)\!e^{-j 2\pi d\frac{s}{S}},
\end{align}
\end{small}
\!\!\!\!wherein $N^{\{2\}}_{\mathrm{cl}}$ and $N^{\{2\}}_{\mathrm{ray}}$ indicate the number of clusters and rays for $h_{d,k}$, while $N^{\{3\}}_{\mathrm{cl}}$ and $N^{\{3\}}_{\mathrm{ray}}$ are the number of clusters and rays for $h_{r,k}$. Note that $\Bar{\phi}_{\textrm{BS},k}^{{l,u}}[s]$ and $\Bar{\phi}_{\textrm{RIS},k}^{{l,u}}[s]$ in (\ref{channel2}) and (\ref{channel3}), respectively, denote the spatial AoD of the BS and the RIS towards the user $k$.
Let $\mathbf{\Theta}=$diag$\left(\theta_{1},\theta
_{2},...,\theta_{M}\right)$ be the diagonal phase shift matrix with $\theta_{m}$, denoting the reflection coefficient of the $m$th element in the RIS. Accordingly, $\theta_{m}=\beta_{m}e^{j\varphi_{m}}$ with $\beta_{m}\in$ $[0,1]$ and $\varphi_{m}\in$ $[0,2\pi]$ representing the amplitude and the step of the $m$-th RIS reflecting element.
Hence, the downlink received signal at the user $k$ over the subcarrier $s$ is represented by
$
y_{k}[s]=\left(  \mathbf{h}_{d,k}[s]+\mathbf{ G}[s]\mathbf{\Theta}\mathbf{h}_{r,k}[s]\right)  x_{k}[s] \mathbf{+}n_{k}[s], \label{R1}%
$
where $x_{k}[s]=\mathbf{w}_{k}[s] c_{k}[s]$ is the transmitted signal at the BS,
$c_{k}[s]$ denotes the transmit data symbol, $\mathbf{w}_{k}[s]\in%
\mathbb{C}^{M\times1}$ indicates the corresponding transmit beamforming vector and finally $n_{k}[s]$ represents the complex additive white Gaussian noise (AWGN) with unit variance all for the $k$th user over the subcarrier $s$. 
\par Let us define the feasible set $\mathcal{F}$ for the RIS reflecting phase shifts as
$
    \mathcal{F}\delequal\{\varphi_{m}\big|~|\varphi_{m}|\le{1}\}~\forall{m\in{M}}
$ that equivalently guarantees
$0\leq\varphi_{m}\leq2\pi,$ $\forall m$\cite{DRL}.
Under the assumption of given beamforming vectors, we seek to optimize 
$\mathbf{\Theta}$ in the feasible set, i.e., $\varphi_m\in{\mathcal{F}},~\forall{m}$, formulated as
\begin{align} \label{prob}
    \mathbf{\Theta}^{*}=\argmax_{\mathbf{\Theta}}
{\textstyle\sum\nolimits_{k=1}^{K}} {\textstyle\sum\nolimits_{s=1}^{S}}
\ln\left(1+\gamma_{k}[s] \right),
\end{align}
where $\gamma_{k}[s]=\dfrac{P}{\sigma^{2}}\Big |\left(  \mathbf{h}_{d,k}[s]+\mathbf{ G}[s]\mathbf{\Theta}\mathbf{h}_{r,k}[s]\right)  \mathbf{w}_{k}[s]\Big|^{2}$ and $P$ denotes the transmit power. This problem is in quadratic form and thus non-convex. While the prior near-optimal optimization-driven solutions (e.g., \cite{9219206}) to (\ref{prob}) cannot support real-time operations, the existing learning-based approaches such as \cite{DRL} and \cite{LSTM} suffer from precision loss especially in large-scale systems. 

\section{Tracking the RIS Reflection Coefficients}
To mitigate the aforementioned drawbacks, we propose a time-series-based deep learning method as a solution to (\ref{prob}), by tracking the historical RIS reflection coefficients and forecasting them at upcoming time steps.
\begin{figure}[tbp]
	\centering \hspace{0.0cm}\includegraphics[width=7.0cm,height=2.8cm]{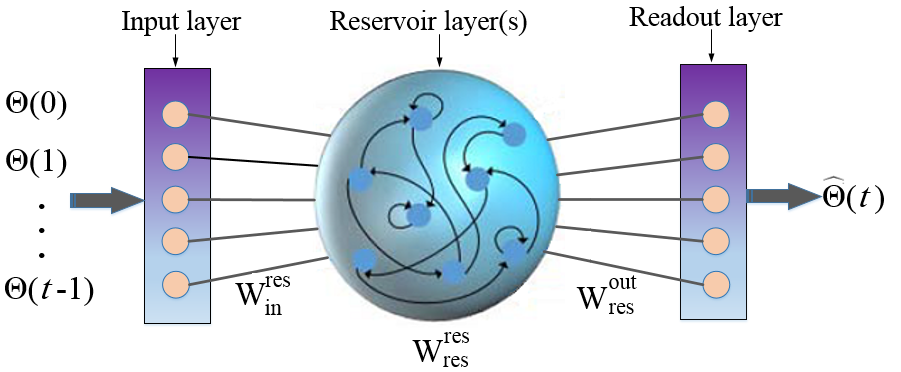}
	\caption{Architecture of LSM.}
	\label{fig:LSM}
	\vspace{-0.50cm}
\end{figure}

\subsection{Time-Series Forecasting}
Advantageous of temporal dynamic behavior and time-varying nature, artificial neural networks (ANNs) can track time-series data (i.e., sequences of data at continuous time steps). Since the RIS reflection coefficients are frequently varying, they can be treated as time-dependent data. In this paper thus, we track the historical variations of the RIS reflection coefficients at preceding time steps based on time-series ANNs to forecast them in future. To this end, we first develop an LSM as a time-series ANN with reservoir layer(s) \cite{schrauwen2007overview} and train it by extreme learning method \cite{huang2006extreme}. Due to the sparse structure of the reservoirs, the LSM can be easily trained
and precisely forecast the future trend. According to Fig.~\ref{fig:LSM}, unlike the fully-connected structure of conventional deep neural networks\cite{LSTM}, reservoir (hidden) layer(s) neurons in the LSM exhibit randomly and sparsely connected structure, leading to a lower training time compared to the conventional time-series ANN.
The input layer of the LSM in this paper is fed through the historical $\mathbf{\Theta}$ i.e., $\boldsymbol{\Theta}(1),\boldsymbol{\Theta}(2),...,%
\boldsymbol{\Theta}(t-1)$. At time step $t$, the reservoir layer state can be updated as \cite{schrauwen2007overview} 

\begin{small}
\begin{align} \label{RC}
	\mathrm{Res}(t)=f^{\mathrm{act}}[\widehat{W}_{\mathrm{res}}^{\mathrm{res}}.\mathrm{Res}(t-1)+\widehat{W}_{\mathrm{in}}^{\mathrm{res}}.{%
		\boldsymbol{\Theta}}(t-1)], 
\end{align}%
\end{small}
\!\!\!\!with $\widehat{W}_{\mathrm{in}}^{\mathrm{res}}$ and $\widehat{W}_{\mathrm{res}}^{\mathrm{res}}$ denoting the input and the reservoir layer, respectively. Note that the reservoir weights $\widehat{W}_{\mathrm{res}}^{\mathrm{res}}$, are randomly
initialized and never tuned throughout the LSM training process. 
\begin{figure}[tbp]
	\centering \includegraphics[width=5.0cm,height=2.8cm]{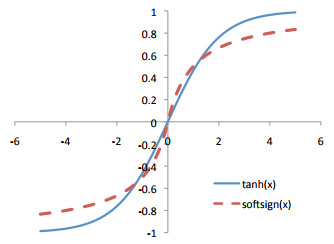}
	\caption{Behaviour of tanh and softsign non-linear functions.}
	\label{fig:act-irs}
	\vspace{-0.50cm}
\end{figure}
In (\ref{RC}), $f^{\mathrm{act}}$ is an activation function such as tanh or softsign functions for non-linearity injection into the training process.
Accordingly, the readout layer forecasts the RIS reflection coefficients at time step $t$ as 
$
	\boldsymbol{\Theta}(t)=\widehat{W}_{\mathrm{res}}^{\mathrm{out}}.\mathrm{Res}(t-1),
$
where $\widehat{W}_{\mathrm{res}}^{\mathrm{out}}$ is the readout weights.
The behaviour of activation functions illuminates the requirement for a proper initialization of the input
weights that consequently guarantees a realistic amount of convergence iterations (epochs). As observed from Fig.~\ref{fig:act-irs}, both functions behave almost linearly for the input values within the zero adjacency.
Obviously, assigning quite small values (near to zero) for the input weights (in x-axis) leads to smaller variance of the input signals in a layer-wise manner according to (\ref{RC}). In this way, the non-linearity feature in $f^{\mathrm{act}}$ and therefore the advantage of training in deep layers will be lost. From another point of view, both functions get flat and saturated, when input signals are of large values. Accordingly, setting too large values for the input weights (in x-axis) leads to a layer-wise increase for the variance of the input signals in  (\ref{RC}) that makes the gradient in Fig.~\ref{fig:act-irs} starts approaching zero. In the next section, we explain how to properly initialize the input signal weights based on Xavier initializer technique~\cite{Xavier}.
\subsection{Xavier Initialization}
We intend to find the variance and the proper distribution, where the
initial values of $\widehat{W}_{\mathrm{in}}^{\mathrm{res}}$ in (\ref{RC}) should be taken from. The Gaussian distribution with zero mean and finite
variance \cite{Xavier} is adopted as a candidate in our work. Let us consider the linear combination  of
$\Theta _{m}\in{\mathbf{\Theta}}$ and $\widehat{w}%
_{\mathrm{in}}^{\mathrm{res}}(m)\in {\widehat{W}_{\mathrm{in}}^{\mathrm{res}}}$, as
$
	y^{\mathrm{lin}}=\widehat{w}_{\mathrm{in}}^{\mathrm{res}}(1).\Theta _{1}+\widehat{w}_{\mathrm{in}}^{\mathrm{res}}(2).\Theta
	_{2}+...+\widehat{w}_{\mathrm{in}}^{\mathrm{res}}({M}).\Theta _{M},
$ for a specific time step and user.
We aim at an identical variance for $y^{\mathrm{lin}}$ and $\Theta _{m}$. The variance of $y^{\mathrm{lin}}$ can be expressed as

\vspace{-0.30cm}
\begin{small}
\begin{align}	\label{var1} 
	\!\!\!\text{Var}(y^{\mathrm{lin}})\!=\! \text{Var}\big(\widehat{w}_{\mathrm{in}}^{\mathrm{res}}(1).\Theta _{1}\!+\!
 \widehat{w}_{\mathrm{in}}^{\mathrm{res}}(2).\Theta _{2}\!+\!...\!+\!\widehat{w}_{\mathrm{in}}^{\mathrm{res}}({M}).\Theta _{M}\big). 
\end{align}%
\end{small}
\!By definition of the variance, we have 
\begin{small}
\begin{align}\label{var2}
	\text{Var}(\widehat{w}_{\mathrm{in}}^{\mathrm{res}}(m).&\Theta _{m})\!= \mathbb{E}(\widehat{w}%
	_{\mathrm{in}}^{\mathrm{res}}(m))^{2}.\text{Var}(\Theta _{m})+\\\nonumber &\mathbb{E}(\Theta _{m})^{2}.\text{Var}(\widehat{w}_{\mathrm{in}}^{\mathrm{res}}(m))+\text{Var}(\widehat{w}_{\mathrm{in}}^{\mathrm{res}}(m)).\text{Var}(\!\Theta _{m}\!),  
\end{align}%
\end{small}

\!\!\!\!\!\!wherein $\mathbb{E}(.)$ is the mean or expectation. According to the Gaussian distribution with zero mean, (\ref{var2}) is expressed as 

$
	\text{Var}(\widehat{w}_{\mathrm{in}}^{\mathrm{res}}(m).\Theta _{m})=\text{Var}(\widehat{w}%
	_{\mathrm{in}}^{\mathrm{res}}(m)).\text{Var}(\Theta _{m}).
$
Thus, (\ref{var1}) can be restated as 
\begin{small}
\begin{align}\label{var3}
	\text{Var}(&y^{\mathrm{lin}}) = \text{Var}(\widehat{w}_{\mathrm{in}}^{\mathrm{res}}(1)).\text{Var}(\Theta
	_{1})+\\\nonumber
	&\text{Var}(\widehat{w}_{\mathrm{in}}^{\mathrm{res}}(2)).\text{Var}(\Theta _{2}) +...+\text{Var}(\widehat{w}_{\mathrm{in}}^{\mathrm{res}}(M)).\text{Var}(\Theta
	_{M}). 
\end{align}%
\end{small}
\par Note that for identical distribution of $\mathbf{\Theta}$ and $\widehat{W}_{\mathrm{in}}^{\mathrm{res}}$, (\ref{var3}) can be written as 
$
	\text{var}(y^{\mathrm{lin}})=M.\text{var}(\widehat{w}_{\mathrm{in}}^{\mathrm{res}}(m)).%
	\text{var}(\Theta _{m}).
$
Therefore, the identical variance for $y^{\mathrm{lin}}$ and $\Theta _{m}$ entails
\begin{align}
	M.\text{var}(\widehat{w}_{\mathrm{in}}^{\mathrm{res}}(m))=1, ~\textrm{and}~
	\text{var}(\widehat{w}_{\mathrm{in}}^{\mathrm{res}}(m))=1/M.
\end{align}
\begin{figure}
\centering\includegraphics[width=9.0cm,height=4.50cm]{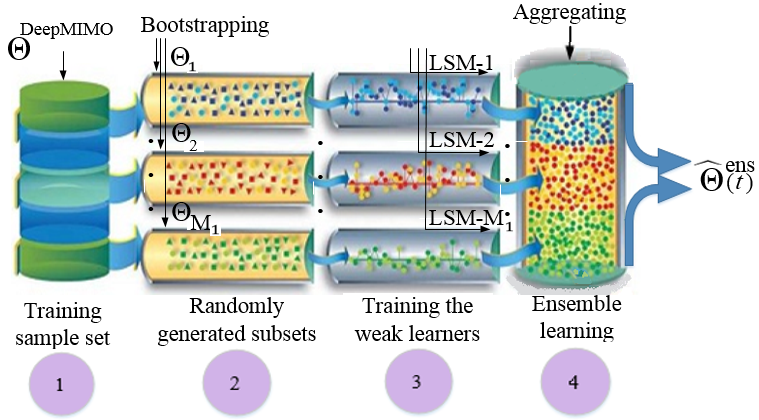}
\caption{Ensemble learning schematic.}
\label{fig:ensemble}
\vspace{-0.50cm}
\end{figure}
Hence, the initial input weights follow a Gaussian distribution
with a zero mean and the finite variance of $1/M$\cite{Xavier}.
\begin{figure*}[t]
	\begin{tabular}{lccccc}
	\centering
		&\hspace{0.750cm}\includegraphics[width=6.0cm,height=4.0cm]{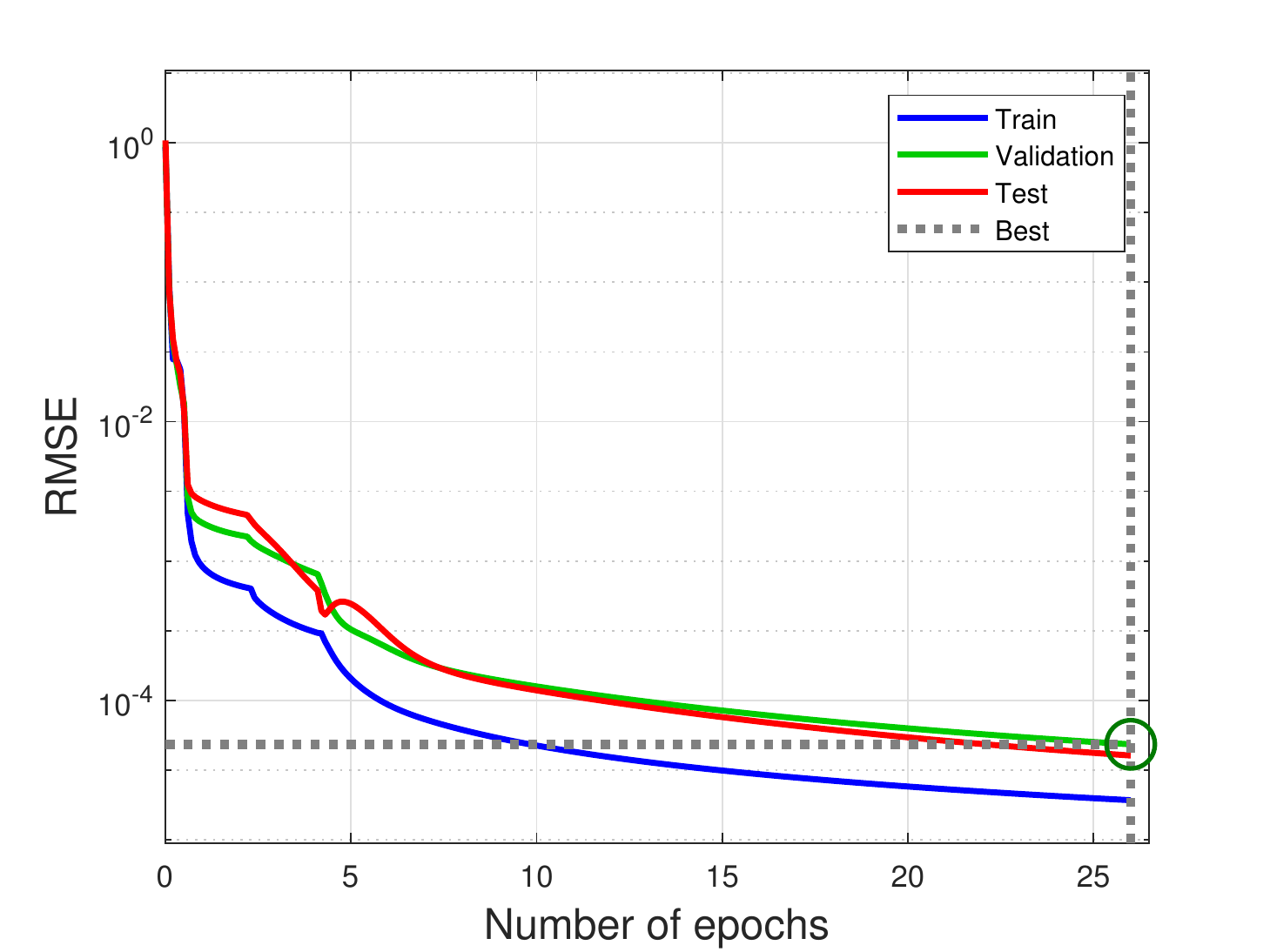}\hspace{3.50cm}\includegraphics[width=5.10cm,height=3.80cm]{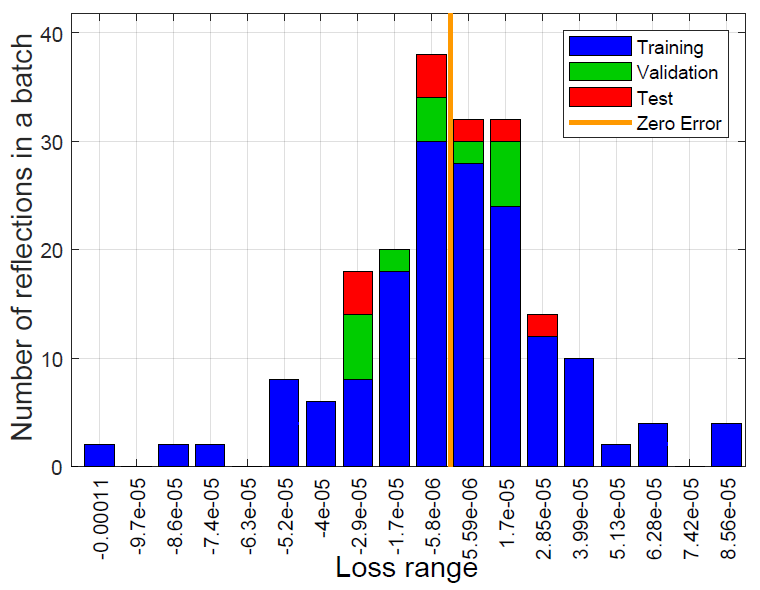}\\
		&\hspace{-7.70cm}(a) RMSE vs. the number of epochs. &\hspace{-6.50cm}(b) The number of batched samples vs. loss range.
			\end{tabular}
	\caption{Performance evaluation of training, validation and test phases of the LSM.} 
	\label{fig:performance1}
	\vspace{-0.30cm}
\end{figure*}
\begin{figure*}[t]
	\begin{tabular}{lccccc}
	\centering
		&\hspace{-0.0cm}\includegraphics[width=6.0cm,height=4.0cm]{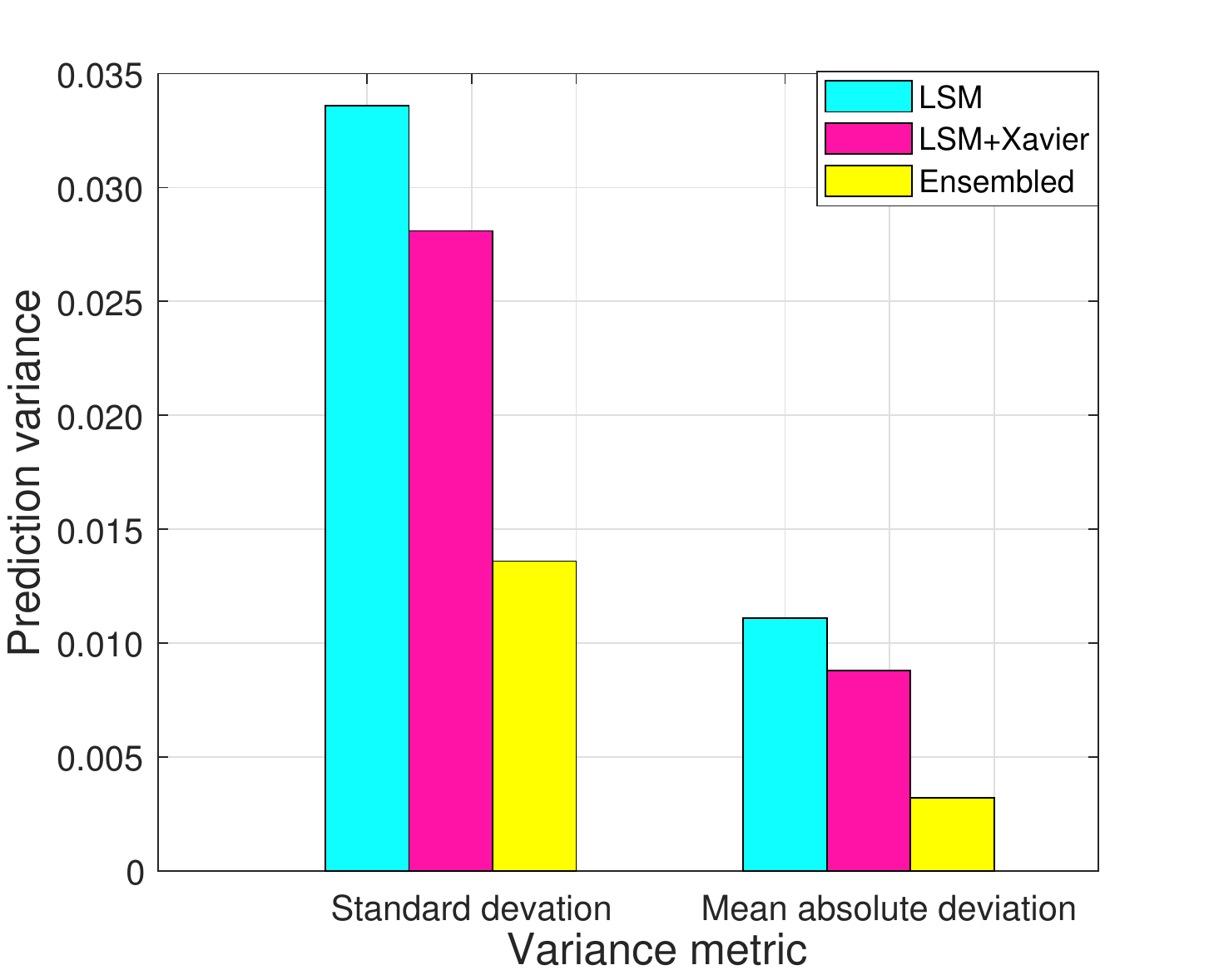}&\hspace{1.3cm}\includegraphics[width=5.80cm,height=4.0cm]{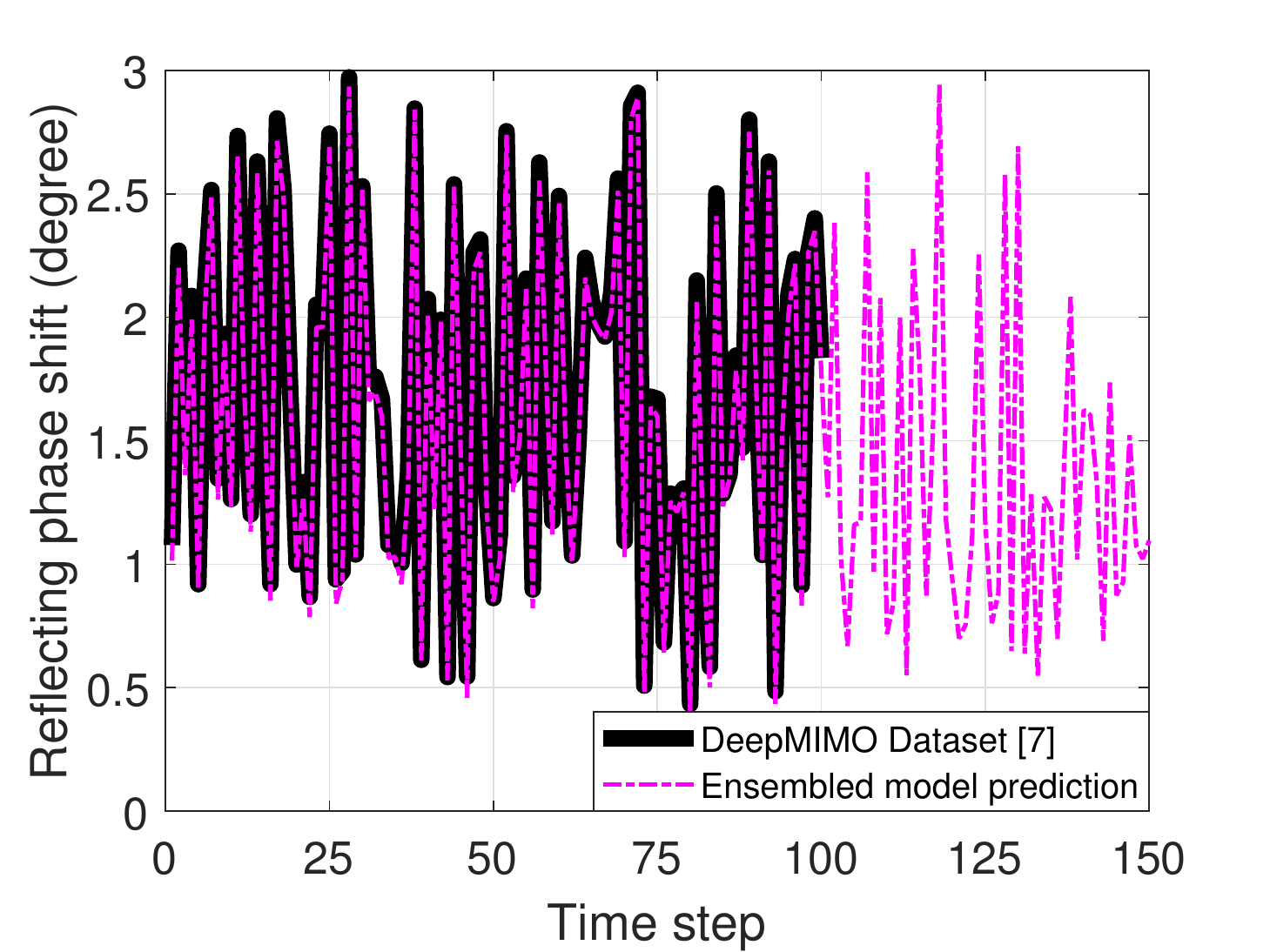}
		\\&\hspace{0.30cm}(a) Analyzing $\tau_1$ and $\tau_2$ for the proposed schemes. &\hspace{1.0cm} (b) RIS reflection coefficients tracking/prediction.		
	\end{tabular}
	\caption{Performance evaluation of the Xavier-enabled LSM and the ensembling model.} 
	\label{fig:performance2}
	\vspace{-0.50cm}
\end{figure*}
\subsection{Ensemble Learning}
As an efficient ensemble learning method, bootstrap aggregation\cite{Bagging} initiates an ensembling model, in which various subsets from the whole training sample dataset are bootstrapped. The LSMs (called weak learners as well) on a different subset are trained in an independent and parallel manner. The well-trained weak learners are capable of predicting the future time steps as shown in Fig.~\ref{fig:ensemble}. By doing so, a more accurate predictor namely strong learner exploits the predictions made by the weak learners through an averaging mechanism for a precise prediction of the future time step. The RIS reflection coefficients tracking in this paper based on bootstrap aggregation (or so-called bagging algorithm) is a regression problem that can be explained as follows. We train $M_1$ weak learners (regressors) through $M_1$ randomly subsets $\Theta_{m}(m\in{M_{1}})$ of the whole training sample set $\mathbf{\Theta}$ obtained from the deepMIMO dataset~\cite{DeepMIMO}. Once the weak learners trained well, they can predict the future time step. Next, the strong learner aggregates the predicted values made by the weak learners and applies a weighted averaging mechanism. The more precise a weak learner is, the more impact it is associated with in final prediction of the strong learner.
Training the weak learners over different subsets is performed in multiple iterations so as to reduce overfitting for the weak learners.

\section{Complexity Analysis}

Since the readout weights are the only ones updated during training the LSM\footnote{The input weights are derived from zero mean finite variance Gaussian distribution through the Xavier initializer technique, while the reservoir weights remain untuned during training the LSM.}, the computational complexity for training the LSM is associated with updating these weights. To this aim, the total number of floating-point operations (FLOPs) metric is invoked in this paper to analyze the computational complexity. Let us denote by $T_{\textrm{max}}$ and $T_{\textrm{0}}$, the maximum training time and initial washout time. Besides, let $N_{\textrm{in}},N_{\textrm{res}}$ and $N_{\textrm{out}}$ denote the number of neurons in input, reservoir and readout layers in LSM, respectively. Accordingly, training the LSM (or equivalently its readout weights $\widehat{W}_{\mathrm{res}}^{\mathrm{out}}$) requires $\mathcal{O}(\textrm{LSM})=(T_{\textrm{max}} - T_{\textrm{0}}+1)^{2}(N_{\textrm{res}}+N_{\textrm{in}}+ T_{\textrm{max}}- T_{\textrm{0}})+(N_{\textrm{res}}+N_{\textrm{in}})(T_{\textrm{max}} - T_{\textrm{0}}+1)(T_{\textrm{max}}-  T_{\textrm{0}})+N_{\textrm{out}}(N_{\textrm{res}}+N_{\textrm{in}})(T_{\textrm{max}} - T_{\textrm{0}})$ addition FLOPs and $(T_{\textrm{max}} -  T_{\textrm{0}}+1)^{2}(N_{\textrm{res}}+2+T_{\textrm{max}}-  T_{\textrm{0}})(N_{\textrm{res}}+N_{\textrm{in}})(T_{\textrm{max}} - T_{\textrm{0}}+1)^{2}+N_{\textrm{out}}(N_{\textrm{res}}+N_{\textrm{in}})(T_{\textrm{max}} - T_{\textrm{0}}+1)^{2}$ multiplication FLOPs\cite{RC1}. Thus, the computational complexity of training the LSM is approximated as $(T_{\textrm{max}} -  T_{\textrm{0}}+1)^{3}$. Additionally, in accordance with simultaneous and independent training of LSMs engaged in ensemble learning, the computational complexity for training the ensembling model is given by $\mathcal{O}(\textrm{LSM}_{\textrm{ens}})=\textrm{max}\{\mathcal{O}(\textrm{LSM}_{1}),\mathcal{O}(\textrm{LSM}_{2}),...,\mathcal{O}(\textrm{LSM}_{M_{1}})\}$.

\section{Simulation Results}
We consider a BS with $N_t=256$ transmit antennas and an RIS with $M=8\times8$ elements for serving $K=4$ users. For the wideband THz communication, the transceivers work at 100 GHz frequency with $S=128$ subcarriers \cite{wideband1}. The number of clusters are $N^{\{1\}}_{\mathrm{cl}}=N^{\{2\}}_{\mathrm{cl}}=N^{\{3\}}_{\mathrm{cl}}=3$, where each cluster includes $N^{\{1\}}_{\mathrm{ray}}=N^{\{2\}}_{\mathrm{ray}}=N^{\{3\}}_{\mathrm{ray}}=1$, respectively. The delays of the clusters and the delay offsets for the rays follow a uniform distribution within $[0, 20]$(ns) and $[-0.1, 0.1]$(ns), respectively. The complex gain also follows a complex Gaussian distribution $\mathcal{CN}(0,1)$\cite{wideband2}. Besides, a fully digital zero-forcing (ZF) beamforming is employed to eliminate inter-user interference. 
For evaluating the LSM training performance, root mean square error (RMSE) metric is invoked as $
\textrm{RMSE} = \sqrt{\sum_{i=1}^{{M}}||\mathbf{\Theta}^{i}(t)-\widehat{\mathbf{\Theta}^{i}}(t)||^{2}_{2}},
$, whereas for evaluating the variance of RIS reflection coefficients tracking  through the LSM, we use mean absolute deviation and standard deviation metrics~\cite{new}, respectively, given by
$
\tau_1 = \frac{\sum_{i=1}^{{Q}}||\widehat{\mathbf{\Theta}^{i}}(t)-\mu||_{2}}%
{Q},
$
and
$
\tau_2 = \sqrt{\frac{\sum_{i=1}^{{Q}}||\widehat{\mathbf{\Theta}^{i}}(t)-\mu||^{2}_{2}}%
{Q}},
$
with $Q$ and $\mu$ denoting the number of predictions and the prediction average, respectively. We observe 100 time slots for tracking the RIS reflection coefficients with 1sec interval in an LSM equipped with 5 reservoir layers and trained by extreme learning (i.e., random weight assignment in reservoir layers). The training and validation steps use 70\% and 30\% of the total samples of the DeepMIMO dataset\cite{DeepMIMO} and we set $M_1$=15 for ensemble learning.
\par According to Fig.~\ref{fig:performance1}(a), training, validation, and test phases of the LSM quickly converges due to the sparsely connected structure of the LSM reservoir layers and employing the extreme learning method (wherein the input and reservoir weights are not trained). Hence, the training is solely limited to readout weights, which is accompanied by negligible RMSE.
\par  Fig.~\ref{fig:performance1}(b) portrays the loss range for training, validation, and test phases when different number of training samples exists within a batch. Clearly, the more number of training samples are provided, the less errors are observed. Another finding in Fig.~\ref{fig:performance1}(b) is that the errors in validation and test phases are mostly experienced adjacent to the zero error line, which is trivial. This observation indicates that the LSM is well-trained and operates well in terms of prediction accuracy.
\par Numerically, it is verified in Fig.~\ref{fig:performance2}(a) that fine-tuning the LSM with Xavier initializer technique and then applying the ensemble learning technique leads to remarkable lower LSM prediction variance. In specific, the former case, i.e., Xavier-enabled LSM achieves up to 17\% and 26\% reduction for $\tau_1$ and $\tau_2$, respectively, compared to training the conventional LSM with random initial input weights. This can be justified that the initial parameter optimization in deep learning techniques not only reduces the training epochs needed to converge (as seen formerly in Fig.~\ref{fig:performance1}(a)), but also improves the validation and test accuracy (as verified in Fig.~\ref{fig:performance1}(b)) and thereby provides immunity against overfitting (unlike the LSTM with non-optimized parameters employed in \cite{LSTM}). The latter case, i.e., the ensembling model according to Fig.~\ref{fig:performance2}(a) degrades the LSM prediction variance up to 52\% and 66\% for $\tau_1$ and $\tau_2$, respectively. The remarkable degradation in $\tau_1$ and $\tau_2$ is stemmed from employing the prediction power of various regressors simultaneously for accurate prediction\cite{Bagging}, which in turn leads to further immunity against overfitting.
\begin{figure*}[t]
	\begin{tabular}{lccccc}
	\centering
		\hspace{1.0cm}\includegraphics[width=6.0cm,height=4.0cm]{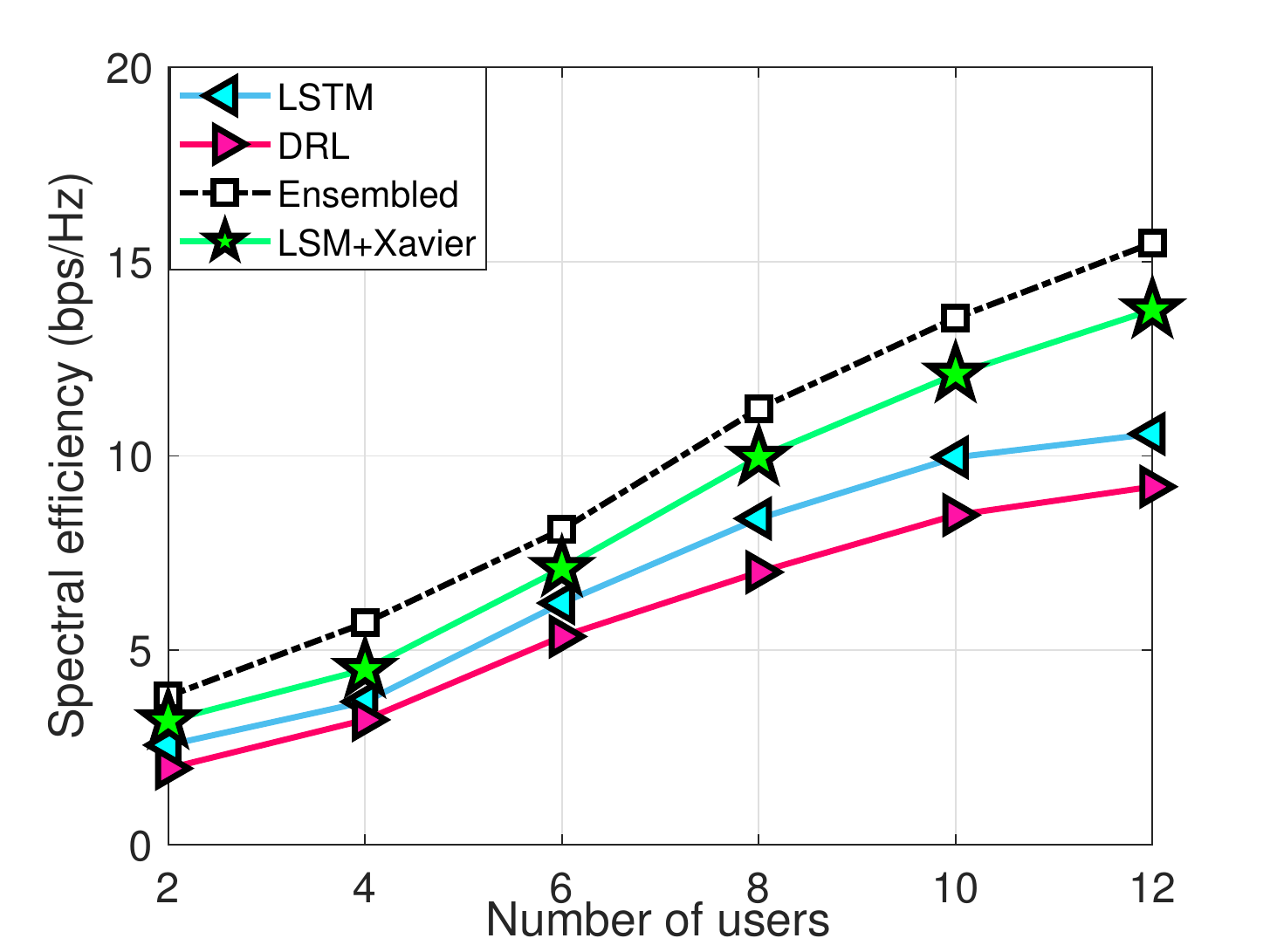}&\hspace{2.50cm}\includegraphics[width=6.0cm,height=4.0cm]{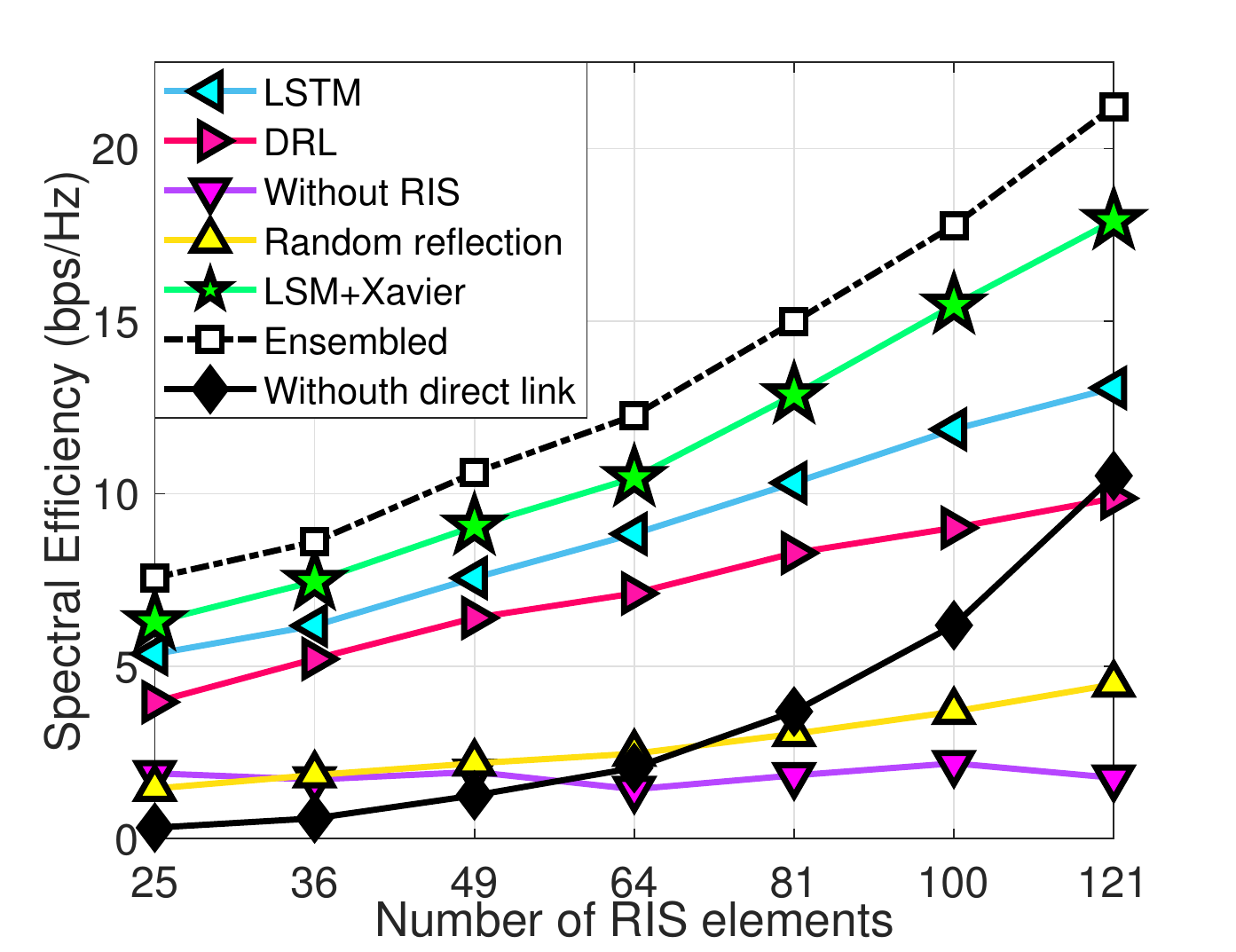}\\
		\hspace{1.0cm}(a) System achievable SE vs. increasing $K$. &\hspace{2.50cm}(b) System achievable SE vs. increasing $||\mathbf{\Theta}||$.		
	\end{tabular}
	\caption{Performance evaluation of the RIS-aided THz system.} 
	\label{fig:performance3}
	\vspace{-0.5cm}
\end{figure*}
\par The excellent precision of the ensembling model and its prediction capability are demonstrated in Fig.~\ref{fig:performance2}(b). The ensembling model, as observed, precisely tracks the phase shift of the reflecting elements in RIS (i.e., $\varphi_m \in{\mathcal{F}}~\forall m$) within first 100 time steps and precisely overlays the given samples in DeepMIMO dataset \cite{DeepMIMO}. Over the upcoming 50 time steps, the ensembling model, as shown, is capable of forecasting the RIS reflection future trend, indicating how the phase shift of the reflecting elements in RIS varies.
\par We analyze the total achievable SE for the proposed RIS-assisted THz system versus increasing the number of users in Fig.~\ref{fig:performance3}(a) compares the proposed Xavier-enabled LSM and ensembling model with DRL\cite{DRL} and LSTM\cite{LSTM} baselines. As observed, our proposed baselines outperform others, thanks to lower prediction variance and thereby higher prediction accuracy (as verified in Fig.~\ref{fig:performance2}(a)). More precisely, the ensembling model and the Xavier-enabled LSM schemes improve the achievable SE for the DRL scheme\cite{DRL} as much as 42\% and 34\%, respectively, when there exist 12 users. Compared to the LSTM scheme \cite{LSTM}, we observe atmost 33\% and 25\% achievable SE gain for the ensembling model and Xavier-enabled LSM schemes, respectively, with 12 users.
\par The simulations are extended by analyzing the total achievable SE versus the number of RIS elements as depicted in Fig.~\ref{fig:performance3}(b). To investigate the role of RIS in enhancing the achievable SE as well as indicating the performance of the proposed RIS reflection coefficients tracking scheme, we account for additional baseline schemes including:
\begin{itemize}
\item \textit{Without~direct~link}: It is assumed that direct links between the BS and the users are all obstructed. Hence, the RIS is the only way of communicating between the BS and the users within the cellular network.
\item \textit{Random~reflection}: It is assumed that the phase shift of the reflecting elements in RIS
are randomly derived from the values in feasible set $\mathcal{F}$. 
\item \textit{Without~RIS}: The conventional cellular network without RIS, wherein the transmissions completely rely on direct links between the BS and the users. 
\end{itemize}
As indicated in Fig.~\ref{fig:performance3}(b), except for ``Without RIS", all the baselines  experience a growth in achievable SE by increasing the number of RIS elements. Evidently, the ``Random reflection" baseline has the mildest growth due to randomized reflecting elements, while our proposed schemes, i.e., the ensembling model and the Xavier-enabled LSM have got the most intensive trend.
Owing to exponential action space in DRL \cite{DRL} and overfitting in LSTM \cite{LSTM} for the large-scale RIS, these two baselines achieve lower SE compared to the proposed baselines in this paper as seen. More specifically in comparison with the LSTM baseline for an RIS of size 11$\times$11, up to 39\% and 28\% more SE is achieved by the ensembling model and the Xavier-enabled LSM baselines, respectively. Under such a system parameter configuration, we also observe at most 54\% and 46\% achievable SE gain compared to the DRL \cite{DRL} baseline for the ensembling model and the Xavier-enabled LSM baselines, respectively. In contrast to the small-scale RIS, it is observed that the baseline ``Without direct link" grows weakly since the communicating signals benefit neither from direct links, nor from effective BS-RIS-user links. For the large-scale RIS nevertheless, this baseline employs an efficient RIS reflecting capability
for penetrating the THz waves, which in turn experiences an improved total achievable SE. Eventually as shown, while the number of RIS elements grows, the gap between the
ensembling model baseline and the Xavier-enabled LSM baseline gets more and more clear. For instance, given an RIS of size 11$\times$11, 17\% more achievable SE is obtained for the ensembling model. This phenomenon mainly originates from incorporating a large number of samples that impose more training errors, which in turn detrimentally influences the prediction capability of the LSM.

\section{Conclusion}
In an RIS-assisted THz wireless communications system, we have proposed a two-step scheme to predict the RIS reflection coefficients. In particular, an LSM is trained to track the historical RIS reflection coefficients and predict them  in next time steps. In the first step, a Xavier initialization technique fine-tunes the initial input weights in LSM and achieves atmost 26\% lower LSM prediction variance and at most 46\% improvement in spectral efficiency for the RIS-aided THz system when compared to the existing methods. In the second step, we have used an ensemble learning technique to exploit the prediction power of simultaneous independent LSMs. Simulations have revealed up to 66\% reduction in LSM prediction variance and  enhancement of the achievable SE by as much as 54\% when compared to state-of-the-art methods.

\end{document}